\documentclass[11pt]{article}

\usepackage[preprint]{acl}

\usepackage{pgfplots}

\usepackage{multirow}
\usepackage[table]{xcolor}

\usepackage{times}
\usepackage{latexsym}
\usepackage{hyperref}
\usepackage{float}

\usepackage[T1]{fontenc}

\usepackage[utf8]{inputenc}

\usepackage{microtype}

\usepackage{inconsolata}

\usepackage{graphicx}

\usepackage[normalem]{ulem}
\usepackage{cleveref}

\definecolor{nvidiagreen}{HTML}{76B900}
\colorlet{nvidiagreen}{nvidiagreen!50!white}

\definecolor{nvidiagreendark}{HTML}{76B900}

\title{A Pocket Offline Model for Simultaneous Speech Translation as CUNI Submission to IWSLT 2026}

\author{Aziz Sharipov Ortega \\
  Charles University, MFF, ÚFAL \\
  \texttt{azdisharipov@gmail.com} \\\And
  Dominik Macháček \\
  Charles University, MFF, ÚFAL \\
  \& University of Edinburgh \\
  \texttt{machacek@ufal.mff.cuni.cz} \\}

\begin{document}
\maketitle
\begin{abstract}

We implement simultaneous translation capability with the offline direct speech-to-text translation model Canary, using the  state-of-the-art policy AlignAtt, and submit it to IWSLT 2026 Simultaneous Speech Translation Shared task for Czech to English and English to German and Italian. 

The strengths of our system are: (1) high translation quality, outperforming similarly sized baselines both in low- and high-latency regimes in computationally unaware simulations; (2) low computational requirements, as the model has only 1B parameters; (3) multilinguality -- support of 25 source and 25 target languages.
\end{abstract}

\section{Introduction}
In this paper, we describe a submission of Charles University (CUNI) to the IWSLT 2026 \citep{adelani-etal-2026-iwslt}
Simultaneous Speech Translation Task \cite{anastasopoulos-etal-2026-findings}. Our system is built on top of Canary-1B-v2\footnote{Throughout this paper, we use a simple term \textit{Canary} to refer to the \textit{Canary-1B-v2} model, although technically Canary is a family of models.} \citep{sekoyan2025canarybv} with the state-of-the-art AlignAtt \citep{papi2023alignatt} simultaneous policy. Following  SimulStreaming \cite{machacek-polak-2025-simultaneous}, the top-performing system in IWSLT 2025 \cite{agostinelli-etal-2025-findings} that used the same approach with Whisper model \cite{radford2022whisper}, we also use Silero Voice Activity
Detection (VAD, \citealp{Silero_VAD}) to reduce noise and hallucination. We validate the latency of our systems in computationally unaware simulation, and also submit a version for computationally aware simulation.

One of the strengths of our system is high multilinguality. The Canary model supports 25 source and 25 target languages for direct translation, which overlaps with three language pairs of IWSLT 2026 that we focus on: Czech to English, English to German, and English to Italian. Additionally, the model has only 1B parameters, making it a strong candidate for pocket-device deployment. Recent work has shown that ASR models of similar scale can be further quantized with minimal accuracy loss on CPU only inference \cite{2604.14493}, suggesting the possibility of running the full simultaneous translation pipeline on constrained 
devices such as smartphones, eliminating a need in an external API and reducing the overall latency.

We follow the approach of repurposing offline speech translation models for simultaneous mode because it has been shown as a simple, effective, and well-performing strategy, e.g.\ in \citet{papi-etal-2022-simultaneous} and in \citet{machacek-polak-2025-simultaneous}. Offline models often offer high quality, multilinguality, and strong general robustness, which frequently outweigh their limitations such as hallucination on partial source sentences, lack of left-only context, or repeated encoding of the whole source buffer with every new incoming chunk. These limitations can be mitigated by dedicated simultaneous methods, but such methods typically require computationally expensive training, e.g.\ for stability on source prefixes \cite{niehues18_interspeech}, for learning to wait or translate \cite{Koshkin2026StreamingTA}, or for building specific simultaneous architectures \cite{arivazhagan-etal-2019-monotonic,kyutai2025hibiki}.
We observe a higher demand for offline speech translation than for simultaneous. New, high-quality offline models are becoming available, yet they are rarely evaluated or deployed in simultaneous mode using the state-of-the-art methods. Canary is one such model, the only previously existing simultaneous implementation \cite{gaido2025simulstream} uses a sliding window approach that is outperformed by AlignAtt. 

Our primary goal is therefore to pioneer Canary usage in simultaneous mode with AlignAtt and evaluate it against state-of-the-art systems, to test whether it can serve as a practical system and as a strong baseline for future research.

We conclude we have reached this goal. The results show competitive and in some cases even superior performance to state-of-the-art baselines while being much smaller in size. Our results show improvements by over 4-5 BLEU points over the organizers' baseline on English to German and Italian and 5-8 BLEU points for Czech to English over the IWSLT 2025 best performing system \citep{machacek-polak-2025-simultaneous}. Our implementation is integrated into the  SimulStreaming project: \url{github.com/ufal/SimulStreaming}.

The paper is structured as follows: in the \Cref{sec:background}, we introduce the core model and policies and provide a relevant background. In \Cref{sec:implementation}, we dive into implementation details of our submission, while \Cref{sec:dev} serves as a detailed description of the frameworks and evaluation metrics used, which we then follow by the evaluation result in \Cref{sec:results}. Finally, we wrap up the paper with a conclusion and limitations regarding the implementation.

\section{Background}
\label{sec:background}
\paragraph{Canary-1B-v2} \citep{sekoyan2025canarybv} is a strong multitask speech transcription and translation model with state-of-the-art performance. It outperforms \textit{whisper-large-v3} \citep{radford2022whisper} and \textit{seamless-m4t-medium} \cite{seamlessm4t} in automatic speech recognition (ASR) and automatic speech translation (AST) tasks, and on some domains and language directions is even superior to s\textit{eamless-m4t-v2-large}, which is twice as large in size. Additionally, it supports injecting context in the decoder prompt to bias the prediction with in-domain context, and provides word-level timestamps. Unlike Whisper, Canary does not need to work with fixed size audio inputs and was instead trained to support audios in 0 to 40 seconds range, which makes it a perfect candidate for simultaneous adaptation.

\paragraph{AlignAtt} \citep{papi2023alignatt} is a simultaneous
policy, a method to process an offline AST or ASR model on output that is incrementally growing, one chunk at a time. The core idea AlignAtt is to use the cross-attention of the decoder to omit the suffix of hypothesis after the first token that attends to a predefined frame threshold. It has been shown to work well in long-form translation \cite{papi-et-al-2024-streamatt}, but has not been previously applied on Canary.

\paragraph{Sliding window} \citep{sen2022simultaneous} is another simultaneous policy that we use as a contrastive baseline. Its core idea is to re-translate a window of audio input, use longest common subsequence with the previous translation to avoid repeating, and slide the window with newly available audio chunk.

\section{Implementation}
\label{sec:implementation}

Let us describe the details of the implementation of our systems and the end-to-end long-form speech-to-text processing pipeline. 

\paragraph{Simultaneous frameworks}
Evaluating simultaneous speech translation typically relies on frameworks that simulate real-time conditions on pre-recorded audio, enabling reproducible benchmarking. Two such frameworks support speech-to-text simultaneous systems and allow integration of new models: SimulStreaming \citep{machacek-polak-2025-simultaneous} and Simulstream \citep{gaido2025simulstream}. We use primarily SimulStreaming because it provides a robust implementation of SileroVAD \cite{Silero_VAD}, 
a streaming voice activity detection that filters non-voice parts of the input audio and allows to save computational power on processing empty input, while also avoiding potential hallucinations.

Since IWSLT 2026 task required Simulstream for computational aware evaluation, we finally transfer our Canary implementation to Simulstream to enable it. However, in case of differences, we consider our SimulStreaming implementation as the primary.

\paragraph{Adapting Nemo} Applying the AlignAtt policy to Canary requires decoder forced prefix injection\footnote{The approach is described in this thread \url{https://github.com/openai/whisper/discussions/117\#discussioncomment-3727051}} for incremental output, which the NeMo API did not natively support. Hence, we highlight our contribution to the Nemo's speech framework \citep{Harper_NeMo_a_toolkit} that allows for this core incremental concept to be used with Canary. With our contribution, the prefix can be provided as an optional initial prompt to the decoder. Additionally, we fixed the bug in the cross-attention outputs for the beam decoder strategy, which allowed us to get deterministic dimension outputs and map the cross-attention scores to their respective output tokens.

\paragraph{Processing loop} Our audio processing pipeline follows the stages of prototypical systems described in \citet{papi-etal-2025-real}, Section 3. We keep the same numbering of steps:

\begin{enumerate}
\item[1.-2.] Audio acquisition and segmentation using VAD is identical to SimulStreaming. Refer to \citet{machacek-polak-2025-simultaneous}, Section 3.

\item[3.] Speech buffer update. The incoming chunk, which has \textit{MinChunkSize}\footnote{We mark system parameters with italics.} seconds if the end of voice is not detected, or less otherwise, is concatenated with the speech buffer.

\item[4.] Hypothesis generation. Canary is an attention based encoder-decoder (AED) model. When audio is inputted, it is first encoded by the encoder. Then an initial prompt is passed to the decoder, which contains information about the source and target languages, decoder forced-prefix if any is in the buffer, as well as some other special tokens that affect final output, but are not relevant in the context of implementation, for example, whether to use timestamps. Then the model decodes the target as long as the AlignAtt policy allows. If any part of a word is inside the AlignAtt's \textit{Frames} threshold, the word is removed from the output. If the current chunk is not final, the decoding continues until the most attended source frame is close to the end of the audio, which is indicated by the \textit{Frames} parameter. In case the current chunk is final, we do not apply AlignAtt and output the whole generated sequence. 

\item[5.] Audio and context buffers. We use the following two buffers: (1) source audio buffer, which accumulates latest 30 seconds of audio in raw form. We note that although storing the mel-features instead of raw form is more efficient, as it skips the additional preprocessing step on every new chunk, we implement a simpler buffer as we were more focused on the computationally unaware scenario, and (2) forced decoding target buffer that contains the stable part of the hypothesis that was decoded from current audio buffer.

If the audio buffer has length of 30 seconds or more, we remove the first speech chunk from the source audio buffer. At the same time, we discard the text that was decoded with the first chunk from forced decoding. After shifting the buffer, it may happen that the audio is not entirely parallel to the forced decoded target buffer, but the results show that the model performs well with that. We leave possible improvement to further work.

\end{enumerate}

\section{Development}
\label{sec:dev}

\def\sswhisper{SimulStreaming~Whisper}
\def\highlat{high}
\def\lowlat{low}
\begin{table*}[ht!]
    \centering
    \renewcommand{\arraystretch}{1.15}
    \footnotesize
    \begin{tabular}{ll|c|rrrr}
        & & \textbf{Reg.} & \textbf{BLEU} & \textbf{chrF} & \textbf{XCOMET-XL} & \textbf{LongYAAL (ms)} \\
        \hline
        \multirow{7}{*}{\textbf{En$\rightarrow$De}}
        & \cellcolor{nvidiagreen}Canary ours & \cellcolor{nvidiagreen}\highlat{} &
          \cellcolor{nvidiagreen}\textbf{31.73} &
          \cellcolor{nvidiagreen}\textbf{60.83} &
          \cellcolor{nvidiagreen}\textbf{0.8776} &
          \cellcolor{nvidiagreen}3761 \\
        & baseline organizers (ctx) & \highlat{} & 27.66 & 59.92 & 0.8428 & \textbf{3353} \\
        & baseline organizers (no ctx) & \highlat{} & 27.44 & 59.66 & 0.8351 & 3431 \\
        & Canary offline & -- & 25.01 & 55.52 & 0.7932 & -- \\
        & Canary sliding window & \highlat{} & 23.65 & 58.53 & 0.7922 & 2925 \\
        & \cellcolor{nvidiagreen}Canary ours & \cellcolor{nvidiagreen}\lowlat{} &
          \cellcolor{nvidiagreen}20.70 &
          \cellcolor{nvidiagreen}52.60 &
          \cellcolor{nvidiagreen}\textbf{0.7744} &
          \cellcolor{nvidiagreen}\textbf{1677} \\
        & baseline organizers (ctx) & \lowlat{} & \textbf{22.59} &\textbf{ 57.51} & 0.7651 & 1747 \\
        \hline
        \multirow{7}{*}{\textbf{En$\rightarrow$It}}
        & \cellcolor{nvidiagreen}Canary ours  & \cellcolor{nvidiagreen}\highlat{} &
          \cellcolor{nvidiagreen}\textbf{43.56} &
          \cellcolor{nvidiagreen}\textbf{68.32} &
          \cellcolor{nvidiagreen}\textbf{0.8227} &
          \cellcolor{nvidiagreen}3282 \\
        & baseline organizers (ctx) & \highlat{} & 37.76 & 65.77 & 0.7877 & \textbf{3231} \\
        & baseline organizers (no ctx) & \highlat{} & 37.28 & 65.44 & 0.7806 & 3300 \\
        & Canary sliding window & \highlat{} & 35.52 & 66.07 & 0.7729 & 2724 \\
        & Canary offline & -- & 36.78 & 62.22 & 0.7054 & -- \\
        & \cellcolor{nvidiagreen}Canary ours  & \cellcolor{nvidiagreen}\lowlat{} &
          \cellcolor{nvidiagreen}\textbf{34.79} &
          \cellcolor{nvidiagreen}62.21 &
          \cellcolor{nvidiagreen}\textbf{0.7618} &
          \cellcolor{nvidiagreen}1972 \\
        & baseline organizers (ctx) & \lowlat{} & 31.45 & \textbf{63.03} & 0.6960 & \textbf{1735} \\
        \hline
        \multirow{4}{*}{\textbf{Cs$\rightarrow$En}}
        & \cellcolor{nvidiagreen}Canary ours & \cellcolor{nvidiagreen}\highlat{} &
          \cellcolor{nvidiagreen}\textbf{32.01} &
          \cellcolor{nvidiagreen}\textbf{59.26} &
          \cellcolor{nvidiagreen}\textbf{0.8133} &
          \cellcolor{nvidiagreen}3641 \\
        & \sswhisper{} & \highlat{} & 24.20 & 50.36 & 0.6995 & \textbf{3512} \\
        & \cellcolor{nvidiagreen}Canary ours & \cellcolor{nvidiagreen}\lowlat{} &
          \cellcolor{nvidiagreen}\textbf{27.78} &
          \cellcolor{nvidiagreen}\textbf{56.68} &
          \cellcolor{nvidiagreen}\textbf{0.7633} &
          \cellcolor{nvidiagreen}1997 \\
        & \sswhisper{} & \lowlat{} & 22.11 & 49.55 & 0.6567 & \textbf{1804} \\
        
    \end{tabular}
    \caption{Dev set results of Canary with AlignAtt for simultaneous translation compared to baselines. Reg.\ indicates latency regime: high $<$~4 seconds, low $<$~2 seconds. LongYAAL is reported in milliseconds; lower is better. Our system is highlighted by green background. Baseline organizers entries differ by use of transcript context (ctx) or not (no ctx). Best individual metric results in each language pair and regime are bolded.}
    \label{tab:dev_compare}
\end{table*}

\paragraph{Dev sets} For English-German and English-Italian directions, we use the MCIF \citep{papi2025mcif} dataset, as provided by the IWSLT 2026 organizers. For Czech-to-English direction, we use the
IWSLT 2026 dev set, which cosists of meetings of the Czech Chamber of Deputies \cite{parczech}.

\paragraph{MT metrics}
In our initial development, we relied on BLEU \cite{papineni-etal-2002-bleu} and ChrF \citep{popovic-2015-chrf}. However, to select the final system candidates, we used COMET-XL \citep{guerreiro-etal-2024-xcomet} as it is the top-performing and primary metric at IWSLT 2026.
\paragraph{Latency metrics}
We use computationally unaware LongYAAL \citep{polk2025better} to select best hyper-parameters in the low- and high-latency ranges (2 and 4 seconds respectively).

\section{Results}
\label{sec:results}

We evaluate our systems on the MCIF dev set for English-to-German and English-to-Italian, and on the IWSLT 2026 dev set for Czech-to-English. We compare against the following baselines: (1) the organizers' cascade baseline, which consists of a local agreement ASR component \textit{Qwen3-ASR-1.7B} \citep{shi2026qwenasr} followed by a neural MT model \textit{Qwen3-4B-Instruct-2507} \citep{yang2025qwen}; (2) the Canary sliding window system as implemented in Simulstream, (3) Canary in offline mode, using the default \texttt{transcribe} function in the Nemo toolkit;
and finally (4) SimulStreaming with the direct Whisper model for Czech to English direction.

\paragraph{Metric scores and comparison}

\Cref{tab:dev_compare} reports our main results for the English-to-German and English-to-Italian directions. Our Canary with AlignAtt system outperforms the organizers' baseline across all four configurations -- high and low latency for both language pairs -- on BLEU, chrF, and XCOMET-XL. The improvements are most pronounced in the high-latency regime, where we gain over 4 BLEU points on English-to-German and more than 6 BLEU points on English-to-Italian. XCOMET-XL scores follow the same trend, with 0.042 gains for both language pairs. Notably, this holds even when comparing against the strongest organizers' baseline configuration that uses transcript context -- our system outperforms it on all quality metrics in the high-latency regime for both language pairs. In the low-latency regime, the quality gains are more modest on BLEU and chrF, especially for English to German direction, where the system falls behind the baseline, but XCOMET-XL consistently favors our system.

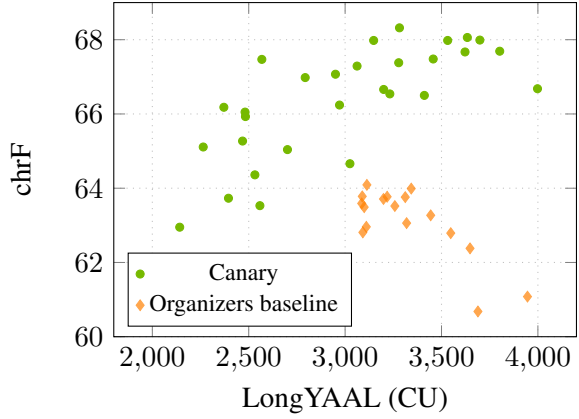
\begin{figure}[t]
    \begin{tikzpicture}
        \begin{axis}[
            width=\columnwidth,
            height=6cm,
            xlabel={LongYAAL (CU)},
            ylabel={chrF},
            xmin=1800, xmax=4200,
            ymin=60, ymax=69,
            legend pos=south west,
            legend style={font=\small, legend columns=1},
            grid=major,
            grid style={dotted},
        ]
        \addplot[only marks, mark=*, mark size=1.5pt, color=nvidiagreendark, opacity=1.0] coordinates {
            (2368,60.68) (2024,60.99) (2162,61.47) (2017,61.72)
            (2142,62.95) (2558,63.53) (2395,63.73)
            (2532,64.36) (3025,64.66) (2701,65.04) (2264,65.11) (2468,65.27)
            (2483,65.93) (2481,66.05) (2371,66.18) (2971,66.24) (3411,66.50)
            (3232,66.54) (3200,66.66) (3998,66.68) (2793,66.98) (2950,67.07)
            (3062,67.29) (3278,67.38) (2568,67.47) (3457,67.48) (3622,67.67)
            (3802,67.69) (3148,67.98) (3532,67.98) (3699,67.99) (3634,68.06)
            (3282,68.32)
        };
        \addlegendentry{Canary}

        \addplot[only marks, mark=diamond*, mark size=2pt, color=orange, opacity=0.7] coordinates {
            (3351,57.83) (3661,59.59) (3689,60.68) (3946,61.08) (3648,62.38)
            (3548,62.79) (3092,62.81) (3110,62.96) (3319,63.06) (3444,63.27)
            (3099,63.49) (3258,63.52) (3086,63.59) (3199,63.71) (3312,63.76)
            (3218,63.77) (3089,63.78) (3343,63.99) (3113,64.09)
        };
        \addlegendentry{Organizers baseline}

        \end{axis}
    \end{tikzpicture}
    \caption{English-to-Italian dev chrF vs.\ LongYAAL (CU) for Canary vs.\ the organizers baseline. Each green point represents one Canary candidate from the search of \textit{MinChunkSize} and \textit{Frame} parameters. Orange points represent the organizers baseline system with grid-search for \textit{segment-length} and \textit{step-length.}} 
    \label{fig:canary-baseline-enit}
\end{figure}

\Cref{fig:canary-baseline-enit} further illustrates the quality-latency trade-off on English-to-Italian. Our Canary system dominates the organizers' cascade baseline across the entire latency range, achieving higher chrF at comparable or lower latency.

\Cref{tab:dev_compare} additionally reports the scores of the Simulstream authors' Canary sliding window implementation, using the configuration they provide as default (chunk 2s, window length 12, matching threshold 0.1). Our AlignAtt-based system substantially outperforms this re-translation approach: on English-to-German, we gain over 8 BLEU points in the high-latency regime, and on English-to-Italian, we gain over 7 BLEU points at a lower latency. This confirms that, while the sliding window policy is a practical and simple way to repurpose Canary for simultaneous mode, AlignAtt provides a clearly superior quality-latency tradeoff.

For the Czech to English direction, we compare our system against the SimulStreaming baseline, which uses AlignAtt with Whisper. As shown in \Cref{tab:dev_compare}, Canary with AlignAtt substantially outperforms this baseline in both latency regimes. In the high-latency setting, our system improves BLEU by nearly 8 points and achieves large gains in chrF as well. In the low-latency regime, we observe a similarly strong improvement of more than 5 BLEU points.

Finally, in \Cref{tab:dev_compare} we report the results of running the model on the dev set in the offline mode. Canary's offline inference on long audios works as described in \citet{sekoyan2025canarybv} in Section 6.4.1. We show that our approach works better than the offline mode and can be used instead if the running time is not a concern.

\paragraph{Grid-search}
We perform grid search to find the optimal \textit{MinChunkSize} and \textit{Frames} parameters to meet the latency thresholds of the IWSLT 2026 Simultaneous task, which is below LongYAAL 2000 ms for low latency, and below LongYAAL 4000ms for high latency, both in computationally unaware simulations. \Cref{tab:canary-csen} showcases the influence of different combinations of the hyper-parameters on the system's performance in Czech to English.

\begin{table}[]
    \centering
    \begin{tabular}{rr|rrr}
Chunk & Frame & BLEU & chrF & Latency \\
\hline
\multicolumn{5}{c}{\textit{Latency $<$ 2s:}} \\
\hline
0.5 & 12 & 27.78 & 56.68 & 1997 \\
2.5 &  1 & 23.86 & 53.72 & 1991 \\
1.0 &  5 & 22.67 & 53.35 & 1652 \\
1.0 &  4 & 20.13 & 51.85 & 1390 \\
1.5 &  4 & 20.07 & 51.83 & 1899 \\
\hline
\multicolumn{5}{c}{\textit{Latency $<$ 4s:}} \\
\hline
2.5 & 20 & 32.01 & 59.14 & 3641 \\
3.5 & 16 & 31.90 & 59.26 & 3921 \\
2.0 & 20 & 31.69 & 59.24 & 3240 \\
3.5 & 12 & 31.43 & 59.01 & 3968 \\
3.0 & 12 & 31.36 & 58.95 & 3954 \\
    \end{tabular}
    \caption{Top-performing Canary Czech-to-English candidates for both latency regimes. We report BLEU (the higher, the better), chrF, and LongYAAL (CU) latency on the Czech-to-English dev set.}
    \label{tab:canary-csen}
\end{table}

\section{Conclusion}

We presented a compact and practical approach to simultaneous speech translation based on the offline Canary-1B-v2 model and the AlignAtt policy. Our main contribution is an end-to-end implementation that adapts a strong offline speech translation model to simultaneous use in the Nemo ecosystem, together with the necessary support for forced-prefix decoding and cross-attention-based truncation.

The results show that this repurposing approach is effective in practice. Across the evaluated language pairs, Canary with AlignAtt achieves competitive quality and in several settings clearly outperforms the organizers' baselines and the previously available Canary sliding-window implementation. At the same time, the system remains lightweight, with only 1B parameters, and supports a relatively broad multilingual setup, which makes it attractive for deployment in resource-constrained scenarios. Our results therefore also serve as a quality reference for what such a lightweight deployment could achieve compared to larger server-side systems and cascade pipelines. We leave the evaluation of quantized Canary in simultaneous mode as a natural next step toward practical edge deployment.

\section*{Limitations}
We also report that during our experiments with the decoder prompt and context, we have not found an optimal setup to bias in-domain predictions. Injecting both forced prefix and context makes the model stall and not produce any output. We presume it could be out of training data.

\section*{Acknowledgements} 
This work was supported by Czech Operational Program OP JAK, the MSCA CZ project MSCA Fellowships -- UK 4, CZ.02.01.01/00/22\_010/0013392, ``LCT''.

\bibliography{iwslt26-bibs,iwslt25-bibs,anth-bibs}

\appendix

\end{document}